\documentclass[acmsmall,screen,nonacm]{acmart}

\usepackage{graphicx}
\usepackage{amsmath}
\usepackage{amsthm}
\usepackage{booktabs}
\usepackage{algorithm}
\usepackage{algorithmic}
\def\mbE{\mathbb{E}}
\newcommand{\AlgName}{\textsc{GTG-Shapley}}

\begin{document}

\title{GTG-Shapley: Efficient and Accurate Participant Contribution Evaluation in Federated Learning}

\author{Zelei Liu}
\affiliation{
    \institution{School of Computer Science and Engineering, Nanyang Technological University}
    \country{Singapore}
}
\email{zelei.liu@ntu.edu.sg}

\author{Yuanyuan Chen}
\affiliation{
    \institution{School of Computer Science and Engineering, Nanyang Technological University}
    \country{Singapore}
}
\email{yuanyuan.chen@ntu.edu.sg}

\author{Han Yu}
\affiliation{
    \institution{School of Computer Science and Engineering, Nanyang Technological University}
    \country{Singapore}
}
\email{han.yu@ntu.edu.sg}

\author{Yang Liu}
\affiliation{
    \institution{Institute for AI Industry Research, Tsinghua University}
    \country{China}
}
\email{liuy03@air.tsinghua.edu.cn}

\author{Lizhen Cui}
\affiliation{
    \institution{School of Software, Shandong University}
    \country{China}
}
\email{clz@sdu.edu.cn}


\begin{abstract}
    Federated Learning (FL) bridges the gap between collaborative machine learning and preserving data privacy. To sustain the long-term operation of an FL ecosystem, it is important to attract high quality data owners with appropriate incentive schemes. As an important building block of such incentive schemes, it is essential to fairly evaluate participants' contribution to the performance of the final FL model without exposing their private data. Shapley Value (SV)-based techniques have been widely adopted to provide fair evaluation of FL participant contributions. However, existing approaches incur significant computation costs, making them difficult to apply in practice. In this paper, we propose the Guided Truncation Gradient Shapley (GTG-Shapley) approach to address this challenge. It reconstructs FL models from gradient updates for SV calculation instead of repeatedly training with different combinations of FL participants. In addition, we design a guided Monte Carlo sampling approach combined with within-round and between-round truncation to further reduce the number of model reconstructions and evaluations required, through extensive experiments under diverse realistic data distribution settings. The results demonstrate that GTG-Shapley can closely approximate actual Shapley values, while significantly increasing computational efficiency compared to the state of the art, especially under non-i.i.d. settings. 
\end{abstract}



\keywords{Federated Learning, Contribution Assessment, Shapley Value}

\maketitle

\section{Introduction}
Recently, with the increasing adoption of AI in many fields, data privacy has become a growing concern for the public. In order to protect user privacy, governments around the globe have proposed various data privacy protection laws, such as the General Data Protection Regulation (GDPR) \cite{GDPR}. Federated Learning (FL) \cite{pmlr-v54-mcmahan17a}, which was proposed by Google, is a new machine learning paradigm which can help AI thrive in the new privacy-aware regulation environment. FL can happen on different scales, from light-weight edge devices to enterprise level data silos \cite{FL2019}.

Since the effectiveness of the final model trained via FL depends on the quality of distributedly owned data, one key issue in FL is how to attract participants with high quality data to join a data federation. Various studies have been conducted on designing proper incentive mechanisms for FL  \cite{Kang-et-al:2019,lyu2020fppdl,Yu2020fair,zeng2020fmore}. In general, they can be divided into three steps: 1) evaluating FL participants' contributions to the final model performance, 2) designing the reward schemes, and 3) distributing the rewards. In order to attract and retain high quality data owners to FL, fairness is important throughout this process. Thus, FL participants' contributions need to be fairly evaluated. Nevertheless, as the FL paradigm does not allow direct access to each participant's local data by others, the contribution evaluation step must be performed without inspecting the original data.

Shapley Value (SV) \cite{shapley1953value} is a classic approach to quantify contribution of individuals within a group under the Cooperative Game Theory. It utilizes each participant's contribution to the utility of all possible subset groups it belongs, and assigns an unique value to each participant. 
Here, we provide an example of Shapley Value calculation. Suppose three people A, B and C work together on a project worth 100 points. Based on a given utility evaluation function $V(\cdot)$, we have $V(A)=50$, $V(B)=50$, $V(C)=10$, $V(A,B)=60$, $V(A,C)=90$, $V(B,C)=90$, and $V(A,B,C)=100$. The Shapley Value for each person is computed as the average of contribution over all possible combinations that person has been involved in, as shown in Table \ref{table:sv-eg}.

\begin{table}[h!]
\caption{A Shapley Value calculation example.}\label{table:sv-eg}
\begin{tabular}{|l|l|l|l|l|l|l|l|}
\hline
  & A-B-C & B-C-A & C-A-B & A-C-B & C-B-A & B-A-C & SV \\ \hline
A & 50    & 10    & 80    & 50    & 10    & 10    & 35 \\ \hline
B & 10    & 50    & 10    & 10    & 80    & 50    & 35 \\ \hline
C & 40    & 40    & 10    & 40    & 10    & 40    & 30 \\ \hline
\end{tabular}

\end{table}


SV is designed to ensure fairness in individual contribution evaluation in a coalition, and only needs the final utility achieved by the coalition for calculation, making it well suited to the requirement under FL of not revealing sensitive information. However, SV has its drawbacks, primarily for being computationally expensive. Directing calculating the SV incurs exponential time complexity with regard to the number of participants. For federated learning involving enterprise level participants (e.g., companies), training once on massive local data silos is already time consuming. As the model size may also be large under such scenarios, aggregating and evaluating the FL model is also time consuming. Hence, the original SV cannot be directly applied in the context of FL.

To leverage SV for measuring FL participant contributions, current research focuses is on improving the efficiency of SV approximation while maintaining accuracy. Existing works in this domain can be divided into two main categories: 1) reducing the number of sub-model evaluations required \cite{ghorbani2019data,jia2019towards,wang2020principled}, and 2) accelerating single round of evaluation \cite{song2019profit,wang2020principled}. To accelerate a single round of evaluation, FL can utilize participants' gradient updates to reconstruct various FL sub-models for instead of re-training each variant of the model from scratch. For reducing the total number of evaluation, existing approaches mostly leverage on Monte-Carlo estimation. However, current approaches are all based on randomly sampled permutation evaluation, which may miss important permutations of FL participants for estimating the SVs.

FL Participants' SVs are estimated with their marginal contributions in the given sequence of joining FL model training. Late comers may show smaller perceived marginal contribution to the FL model performance \cite{FL2019}. This can lead to the SV estimation incorrectly converging. For example, a participant may always appear at the bottom of the sampled permutations. Or, in a more common scenario, a participant may always appear behind certain other participants. This issue undermines the fairness of the SV estimation results, as SVs are in theory supposed to eliminate the ordering effect when evaluating participants' contributions. 


To further enhance the efficiency of SV-based FL participant contribution evaluation, we propose the Guided Truncation Gradient Shapley (GTG-Shapley) approach. It does not require extra learning tasks from participants to be performed when evaluating their Shapley value-based contributions. It utilizes previous gradient updates from all participants to reconstruct FL sub-models instead of training them from scratch. During FL sub-model evaluation, GTG-Shapley reduces the need for model reconstruction through guided Monte Carlo sampling based on the importance of FL participant permutations. Through these mechanisms, GTG-Shapley significantly improves the efficiency of computing Shapley values, while closely approximating the actual Shapley values. 

Extensive experiments have been conducted to compare our proposed GTG-Shapley with the state of the art under various FL settings involving both i.i.d. and non-i.i.d. data silos. The results demonstrate that GTG-Shapley significantly outperforms existing approaches in terms of the efficiency and accuracy of approximating Shapley values. It can enhance the capability of FL systems to fairly assess participant contributions at scale, which is important for supporting the advancement of FL incentive mechanism design.

In summary, our contributions in this paper are as follow:
\begin{itemize}
    \item We propose an efficient Shapley Value approximation algorithm for Horizontal Federated Learning, GTG-Shapley, which reconstructs FL models from gradient updates for FL client contribution evaluation with significantly reduced computational costs. 
    \item We further propose a guided Monte Carlo sampling technique including within-round and between-round truncation, to further enhance the efficiency of GTG-Shapley.
    \item Extensive experiments under various FL data settings, including both i.i.d. and non-i.i.d. data distributions, demonstrate that GTG-Shapley not only significantly improves computation efficiency but also achieves high accuracy compared to the actual Shapley values. 
\end{itemize}

\section{Related Work}
To sustain a long-term operation of an FL ecosystem, participants are more willing to join in if they receive proper incentives. Participants are motivated to provide high quality contribution if the reward is based on their contribution evaluation. It is challenging for measuring participants contribution under FL paradigm since direct access to data is prohibited to protect data privacy. 
Existing approaches can be divided into four categories: 1) self-reporting, 2) individual performance, 3) utility game, and 4) Shapley Value (SV)-based methods.


\subsection{FL Participant Contribution Evaluation via Self-Reporting}
Under this category of approaches, FL participants' contributions are measured by their self-reported information regarding their sensitive local data (e.g., data quantity, quality, committed computational and communication resources). 

In \cite{Yu2020fair}, the authors proposed an incentive mechanism to sustain FL long-term operation which compensates participants for their contributions and cost for join in the federation. The contributions are measured based on self-reported data quantity and quality. In \cite{zeng2020fmore}, the proposed incentive scheme utilizes self-reported local data size and the bandwidth between the participants and the federation for calculating utility with Cobb-Douglas function. In \cite{zhang2020hffl}, authors use publicly verifiable factors about the participants (e.g., data quality, data volume, cost of data collection) to assign types to them and then, determine their contribution levels. In \cite{ding2020incentive}, the authors proposed a contract-based FL incentive system in which FL participants can determine their contribution types based on own data usage cost and communication time. In \cite{feng2019joint}, self-reported training data sizes are treated as the contribution values under the assumption that more training data will lead to better FL model performance. In \cite{Tang2021}, the authors proposed an FL incentive schemes that allows the participants to self-report the resources they are willing to commit to the FL task and the expected reward through negotiation. In \cite{Kang-et-al:2019}, the proposed approach also requires participant to self-report local information, however, this is used for user selection instead of quantifying contributions.

The advantage of self-reporting is that it makes contribution evaluation straight-forward. However there is a strong assumption that the self-reported information is truthful. 


\subsection{Individual FL Participant Contribution Evaluation}
The general idea of individual performance-based FL contribution evaluation is that a participant shall be assigned a contribution value for his performance on specific tasks. This category of contribution evaluation approaches focus on individual performance instead of the overall performance of the FL model. 

In \cite{zhao2021crowdsensing}, the authors measure individual contribution based on similarity between local model updates and the aggregated FL model. The assumption is that local models similar to the FL model are more valuable. In \cite{pandey_crowdsourcing_2020}, the contribution is evaluated through local relative accuracy levels for individual participants. 
Individual evaluation can also be performed by each participant. Participants exchange necessary information between each other for rating other participants with locally owned data. In \cite{lyu2020fppdl}, the proposed approach performs pairwise similarity comparison to allow participants to perform mutual evaluation on each others' generated samples. The approach in \cite{chen2020focus} measures individual contributions by calculating the mutual cross-entropy loss between each local model with the aggregated FL model. There are also techniques leveraging on reputation systems \cite{Yu-et-al:2010,Yu-et-al:2013}. In \cite{lyu2020collaborative}, the approach utilizes reputation to evaluate FL participant contribution. A participant's stand alone model accuracy is used to update his reputation.

These approaches generally assume that participants with local models similar to the aggregated model (or with the local models of other participants) are more valuable to the federation (or the other participants concerned). However, this assumption might not always stand, especially in FL settings. As non-i.i.d scenarios are common in FL \cite{Tan-et-al:2021}, for individual participant holds partial knowledge of the learning task, others possessing complementary knowledge tend to be more valuable than those possessing similar knowledge. 


\subsection{FL Participant Contribution Evaluation via Utility Game}
Utility game-based FL participant contribution evaluation focuses on the changes in the coalition utility when a data owner joins. In a utility game \cite{gollapudi2017profit}, a set of players join teams to produce social utility, and receive individual utility in the form of payments in return. 

There are three profit-sharing principles: 1) egalitarian, 2) marginal gain-based, and 3) marginal loss-based. The most common profit-sharing schemes are fair value game, labor union and Shapley Value-based game \cite{augustine2015dynamics}. 
Fair value treats a player's utility as the marginal loss in the overall utility when a player leaves a coalition. Labor union treats a player's utility as the marginal gain in the overall utility when a player joins a coalition. The order of joining a coalition affects the individual utility values in the labor union scheme. Shapley Value utilize the marginal loss to calculate individual utility free of the order of joining the coalition. 

There are FL participant contribution evaluation approaches based on utility games. In \cite{wang2019measure}, the authors proposed a data influence approximation method for horizontal FL using the marginal loss value. In \cite{ghorbani2019data}, the authors proposed gradient Shapley to estimate data values. Gradient Shapley uses the marginal loss of every data point to approximate the Shapley value of the dataset. The outcome is still affected by the order of each data point in each permutation of evaluation order. In \cite{boi2019rewarding}, the authors proposed influence-based FL contribution evaluation. The marginal loss related to the loss function is used as the utility. The purpose of such a design is to incentivize high quality data owners to join FL as early as possible. In \cite{nishio2020estimation}, FL contribution is evaluated by gradient-based fair value scheme. However, for utility game-based approaches, participants' contributions are affected by the order in which they join in the federation. Therefore, SV-based approaches have been proposed to achieve fairer FL participant contribution evaluation.

\subsection{FL Participant Contribution Evaluation via Shapley Value}
SV-based FL participant contribution evaluation has been the main focus of recent research. Introduced in 1953 as a solution concept in cooperative game theory, SV can fairly calculate a player's contribution in a coalition \cite{shapley1953value}.

The original SV calculation incurs exponential computational costs. There have been techniques for improving SV calculation efficiency. In \cite{castro2009polynomial}, random sampling Monte-Carlo (MC) estimation method is proposed to approximate SV with fewer utility evaluations. In \cite{van2018new}, the authors proposed a structured random sampling MC estimation method to approximate SV efficiently. Since the utility values of different coalitions vary without a clear pattern, the proposed method swaps players to different positions in the coalition ordering. Similar approaches have been developed to improve SV estimation efficiency in FL data valuation, which can be applied in FL participant contribution evaluation. In \cite{jia2019towards}, the authors proposed to use group testing to accelerate SV estimation. In \cite{ghorbani2019data}, the authors leveraged SV to quantify the contribution of individual data points to a learning task. Furthermore, they proposed an efficient approach, TMC-Shapley, to estimate SV by using random sampling Monte-Carlo to estimate and truncate unnecessary utility evaluations in each permutation sampling round.

These methods focus on reducing the number of permutations sampled to calculate SVs. However, under each FL participant permutation, the corresponding FL model needs to be trained from scratch. This requires FL participants to incur significant computation and communication overhead, making such approaches not suitable for practical applications.


\subsection{FL Participant Contribution Evaluation via Gradient Shapley Techniques}
Gradient-based Shapley is proposed to avoid redundant retraining of FL models when evaluating SVs. 
Instead of training from scratch, the FL sub-models corresponding to different participant permutations are approximated by gradient updates from the participants. Participants' gradient update in each round are stored in the FL server and used to reconstruct the corresponding FL sub-model when necessary. This approximation technique has been adopted by state-of-the-art SV-based FL participant contribution evaluation approaches. 

In \cite{song2019profit}, the authors proposed two gradient-based SV schemes: 1) one-round (OR) evaluation and 2) multi-round (MR) evaluation. Both schemes use gradient updates to reconstruct sub-models to replace the actual FL model training process. The difference is that MR calculates SVs during every round of FL model training, while OR calculates SV only once after all the training rounds are complete. 
In \cite{wei2020efficient}, the authors proposed the Truncated MR (TMR) approach, which extends MR by eliminating the entire rounds with decay factor $\lambda$ values lower than a pre-defined threshold. These three schemes significantly improves the efficiency of estimating SVs. However, they still require evaluating every possible FL sub-model as in the original SV setting. In \cite{wang2020principled}, the authors proposed the Federated SV approach with two acceleration techniques for within-round SV estimation: 1) random permutation sampling and 2) group testing. However, the number of random sampling is fixed and not tailored to each permutation. This can result in under-sampling or over-sampling, missing important FL participant permutations and causing inaccurate estimation of SVs.

To improve SV estimation accuracy and reduce computational cost, the SV approximation methods should consider: 1) accelerating utility evaluation, 2) the importance of different FL training rounds and their effects on the SVs, and 3) reducing unnecessary evaluations. None of the existing approaches cover all three aspects.
In this paper, the proposed GTG-Shapley approach bridges this gap and simultaneously enhances efficiency and accuracy.

\section{Preliminaries}
\begin{figure}[t]
    \centering
    \includegraphics[width=1\columnwidth]{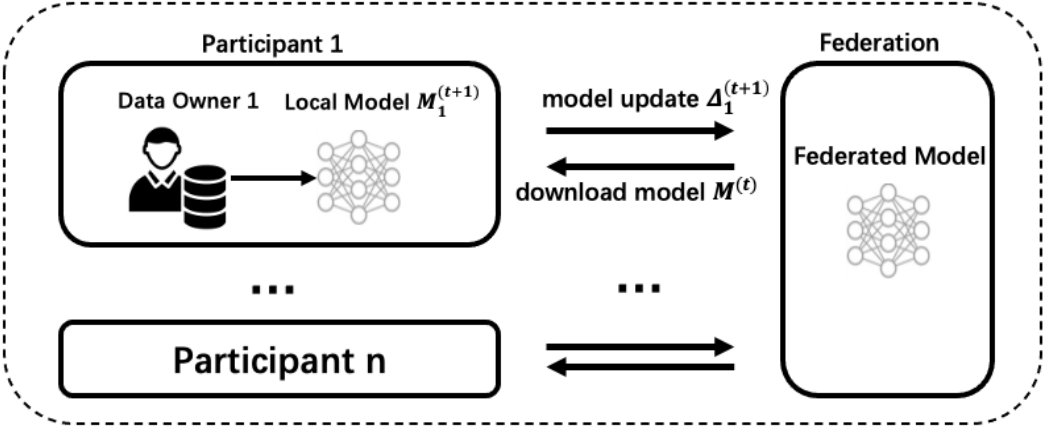}
    \caption{A general horizontal federated learning framework.}
    \label{fig:FL}
\end{figure}

We consider the common Horizontal Federated Learning (HFL) framework first proposed in \cite{FL2019} (as shown in Figure \ref{fig:FL}), in which FL datasets share the same feature space but possess different samples. 

Suppose there are $N=\{1,...,n\}$ data owners, each with a private local dataset $D_i,i\in\{1,\dots,n\}$, who can participate in federated model training. There are $T$ global rounds to train the FL model. During each round $t\in\{1,...,T\}$, participant $i$ downloads the global model $M^{(t)}$, and trains on local dataset $D_i$ for multiple local epochs. Then, participant $i$ obtains a local model $M^{(t+1)}_i$ and sends the gradient update $\Delta^{(t+1)}_i$ to the FL server:
\begin{equation}
    \Delta^{(t+1)}_i=M^{(t+1)}_i-M^{(t)}.
\end{equation}
The FL server gathers participants' gradient updates and executes an aggregation approach, such as FedAvg \cite{pmlr-v54-mcmahan17a} in which  $M^{(t+1)}=\text{FedAvg}(M^{(t)} , \{ \Delta^{(t+1)}_i,i=1,\dots,n \})$, to update the FL model:
\begin{equation}
    M^{(t+1)}=M^{(t)} + \sum_i{\frac{|D_i|}{|D_N|}\Delta^{(t+1)}_i}.
\end{equation}
$|D_i|$ denotes the size of a dataset $D_i$. $|D_N|=\sum_{i=1}^n{|D_i|}$ denotes the size of all $N$ datasets.

Data Shapley value \cite{ghorbani2019data}, $\phi_i(N,V)$, can be used to evaluate the contribution by each participant. It is defined as:
\begin{equation}
\small
    \phi_i(N,V) = \sum_{S \subseteq N\backslash\{i\}}\frac{V(S\cup\{i\}) - V(S)}{\tbinom{|N|-1}{|S|}}.
\label{eq:sv-og}
\end{equation}
$S$ denotes the subset of participants from $N$. The utility function $V(\cdot)$ can assume any form which can evaluate the joint utility of the input.

The Monte-Carlo estimation method for approximating Shapley values from \cite{ghorbani2019data} is adopted:
\begin{equation}
    \phi_i = \mbE_{\pi\sim\Pi}[V(S^i_{\pi} \cup \{i\}) - V(S^i_{\pi})]
\label{eq:sv-estimate}
\end{equation}
where $S^i_{\pi}$ is the set of data from FL participants joining before $i$ in the $\pi$-th counterfactual permutation of the sequence of FL participants. In machine learning, the utility evaluation function $V(S)$ is based on the prediction performance of the model learned using $S$ on a separate test set. That is, $V(S) = V(M_S)$, where $M_S$ is the FL model trained with the subset of datasets $D_S={D_i,i\in S}$ from scratch each time when the utility of $S$ is required:
\begin{equation}
\small
\label{eq:SV-train}
    V(S)=V(M_S)=V(\mathcal{A}(M^{(0)},D_S))
\end{equation}
where $\mathcal{A}$ is the learning algorithm and $M^{(0)}$ denotes the initial model.


In federated learning, model parameters from each participants are collected and aggregated in each iteration by the FL server. Evaluating participants' contributions under the canonical SV settings requires retraining the FL model for each combination of participants \cite{shapley1953value,castro2009polynomial,ghorbani2019data}, which is computationally prohibitive. Replacing FL sub-model evaluation to gradient-based FL model reconstruction \cite{song2019profit,wang2020principled} is necessary.

\begin{figure}[t]
    \centering
    \includegraphics[width=1\columnwidth]{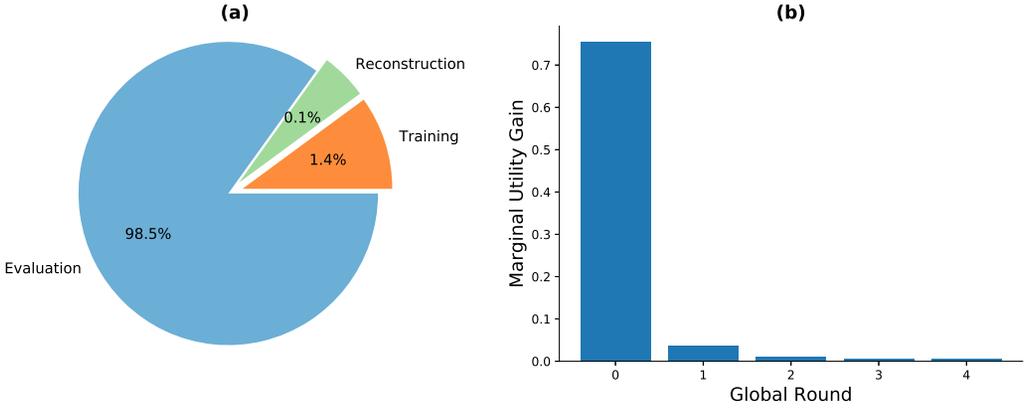}
    \caption{Empirical insights on Gradient-based SV calculation: (a) the distribution of runtime during gradient-based SV calculation; and (b) marginal utility gain in each round of global FL model training.}
    \label{fig:g-SV}
\end{figure}
\begin{figure}[t]
    \centering
    \includegraphics[width=0.7\columnwidth]{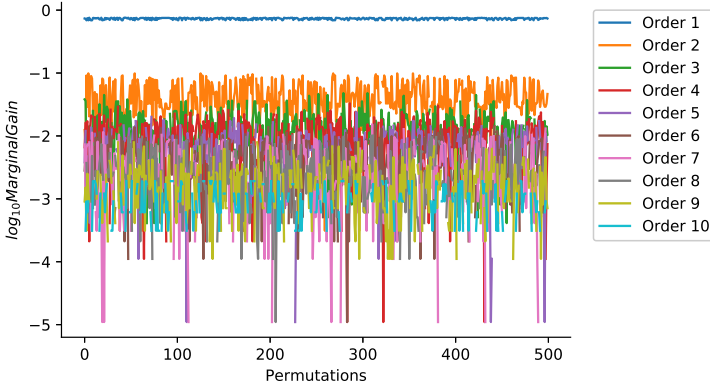}
    \caption{Perceived contributions by participants in terms of marginal utility gain under different orders of joining FL.}
    \label{fig:marginal}
\end{figure}

We analyze the MR approach \cite{wei2020efficient}, which combines the canonical SV calculation and gradient-based model reconstruction. We adopt the non-i.i.d FL data silos setting as described the experiment section with 10 participants and MNIST dataset \cite{lecun2010mnist}. Through such empirical analysis, we derive the following insights. Firstly, the majority of the time required to computing SVs is consumed by FL sub-model performance evaluation (around 98.5\% as shown in Figure \ref{fig:g-SV}(a)). Secondly, the marginal gain in utility (which is defined as model accuracy) between different rounds of global FL model training is significant (as shown in Figure \ref{fig:g-SV}(b)). However, the marginal gain distribution may differ due to the learning tasks, data distributions, or new participants joining. Thirdly, we analyze the effect of participants joining in different orders on the marginal utility (Figure \ref{fig:marginal}). Early participants in every permutations (e.g., Order 1) always contribute large perceived marginal utility, especially the first FL participant. The late joining participants in each permutation always appear less significant. 


\section{The Proposed Approach}
\begin{figure*}[!t]
    \centering
    \includegraphics[width=\linewidth]{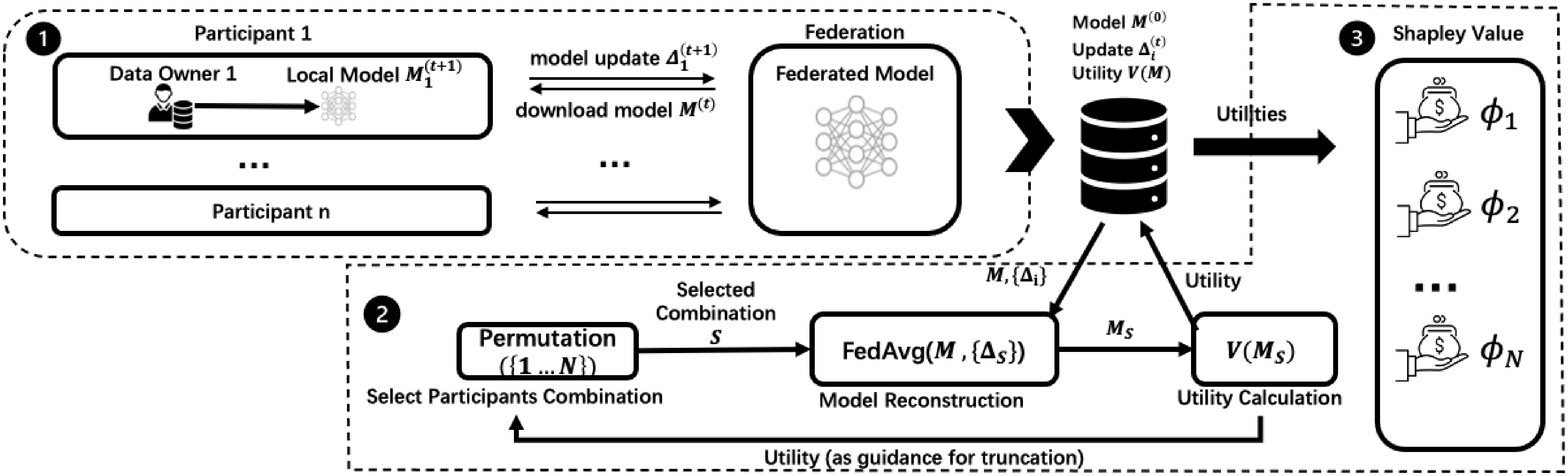}
    \caption{The \AlgName\hspace{1pt}Framework}
    \label{fig:GTG}
\end{figure*}

Based on the aforementioned three insights, we propose the Guided Truncation Gradient Shapley (GTG-Shapley) approach to further enhance the efficiency of contribution evaluation under horizontal federated learning settings \cite{FL2019}. The framework of the proposed approach is illustrated in Figure \ref{fig:GTG}. The key idea is to opportunistically eliminate sub-model retraining and truncate unnecessary model evaluations in order to reducing the computational costs while maintaining the accuracy of estimating SVs.

Under GTG-Shapley, participants join in a federation and train the FL model as they normally do in horizontal federated learning scenarios. However, the gradient updates from each participant during each round of training are stored by the FL server. GTG-Shapley utilizes this information to evaluate the performance of different FL sub-models as a result of different counterfactual permutations of model aggregation based on Shapley value estimation. During sub-model evaluation, GTG-Shapley strategically generates permutations sequences for quick convergence and assesses the necessity of evaluating a sub-model based on its expected marginal gain, and dynamically eliminates those which are insignificant. 

\subsection{Elimination of Sub-Model Retraining}
GTG-Shapley utilizes the gradient updates, $(\Delta)$, as surrogates for datasets to evaluate FL participants' contributions. Thus, Eq. \eqref{eq:SV-train} is re-expressed as:


\begin{equation}
    V(S) = V(M_S) = V\left(M + \sum_{i\in S}{\frac{|D_i|}{|D_S|}\Delta_i}\right).
\end{equation}
In this way, when the calculation of SV requires the evaluation of the utility of $S$, the process of retraining a sub-model $M_S$ is replaced by the process of sub-model reconstruction which is based on previous gradient updates and the current FL model. Hence, the main bottleneck of applying SV in FL is resolved, and SV calculation can be performed entirely on the FL server, thereby incurring no additional computational cost for the participants.

\subsection{Guided Truncation of Model Evaluation} 

With the efficiency of sub-model retraining improved, the majority of the time required to estimate SV is consumed by the model reconstruction and performance evaluation steps (as shown in Figure \ref{fig:g-SV}(a)). For models with a large number of parameters and federations with large test datasets, model evaluation severely slows down SV estimation. Thus, it would be beneficial to eliminate unnecessary model reconstruction and evaluation when estimating SVs following Eq. \eqref{eq:sv-estimate}. 


A participant's Shapley value is estimated based on its marginal gain in different participant permutations, and the FL server performs SV utility evaluation. Based on the insight of utility distribution between different rounds (Figure \ref{fig:g-SV}(b)), the rounds of SV calculation with low marginal utility gain can be discarded, according to Eq. \eqref{eq:sv-estimate}, without significantly affecting SV estimation. Moreover, based on the insight of marginal utility gain contributed by participants in different positions in a permutation (Figure \ref{fig:marginal}), those in later positions contribute little marginal utility gain. Therefore, they can be omitted without significantly affecting SV estimation. 
In addition, this insight also shows that the marginal utility gains concentrate at the front positions in a permutation. Thus, it is important to ensure that participants have fair opportunities to occupy different positions across multiple permutations.

To leverage on these insights, we propose a between-round truncation policy in GTG-Shapley to opportunistically drop entire rounds of SV calculation when the remaining marginal utility gain is small. In addition, a within-round truncation policy is proposed for opportunistically terminating the ongoing sub-model evaluation in permutations when the remaining marginal gain is small. We also propose a guided sampling policy to fairly distribute FL participants in different positions across multiple permutations.

\subsection{The \AlgName\hspace{1pt}Algorithm} 

\begin{algorithm}[b]
\caption{\AlgName}
\label{alg:gtg}
\begin{flushleft}
    \textbf{Input}: initial FL model $M^{(t)}$, final FL model $M^{(t+1)}$, evaluation function $V$, participants' gradient updates $\{\Delta_i,\dots,\Delta_n\}$ \newline
    \textbf{Output}: SVs for round $(t+1)$, $\phi_i^{(t+1)}$, for all $\{i\in\{1,\dots,n\}$ participants
\end{flushleft}
\begin{algorithmic}[1] 
\STATE $\phi_i^{(t+1)}=0, \forall i\in\{1,\dots,n\}$;
\STATE $k=0,v_0=V(M^{(t)}),v_N=V(M^{(t+1)})$;
\STATE \textit{\# between round truncation};
\IF{$|v_N-v_0|>\epsilon_b$}
\WHILE{Convergence criteria not met}
    \STATE $k=k+1$;
    \STATE $\pi^k$: Partial $(n-m)$ permutation of participants; \textit{guided sampling}
    \STATE $v_0^k=v_0$;
    \STATE \textit{\# within-round truncation };
    \FOR{$j=1,\dots,n$}
        \IF{$|v_N-v^k_{j-1}| \ge \epsilon_i $}
            \STATE $C=\{ \pi^k[1],\dots,\pi^k[j] \}$;
            \STATE $\widetilde{M}_C^{(t+1)}=$FedAvg$(\{\Delta_C\},M^{(t)})$;
            \STATE $v^k_j=V(\widetilde{M}_C^{(t+1)})$;
        \ELSE
            \STATE $v^k_j=v^k_{j-1}$;
        \ENDIF
        \STATE $\phi_{\pi^k[j]}^{(t+1)} = \frac{k-1}{k}\phi_{\pi^k[j]}^{(t+1)} + \frac{1}{k}(v_j^k-v_{j-1}^k)$;
    \ENDFOR
\ENDWHILE
\ENDIF
\RETURN $\{\phi_{1}^{(t+1)},\dots,\phi_{n}^{(t+1)}\}$;
\end{algorithmic}
\end{algorithm}

The pseudo code of the proposed GTG-Shapley approach is shown in Algorithm \ref{alg:gtg}.
The initial FL model $M^{(t)}$ is an empty set ($S=\{\empty\}$) for the $t$-th round of SV calculation. The final model $M^{(t+1)}$ is the FL model learnt during the $t$-th round of gradient updating, which is the full set of coalition ($S=\{1,...,n\}$). Lines 1-2 show parameter initialization and the performance evaluation of two input models. GTG-Shapley performs Monte-Carlo truncation at two levels. It performs between-round truncation first. If the marginal gain of the $t$-th round $|v_N-v_0|$ is not larger than a pre-defined threshold $\epsilon_b$, entire round $t$ is ignored, and GTG-Shapley returns 0 for every participant. Otherwise, GTG-Shapley estimates the $t$-th round's SVs. Line 7 is an essential part of GTG-Shapley. The partial permutation is the proposed guided sampling policy with the following rule. The first $m$ ($m<<n$) bits in the sequence are circulated in a fixed order $P(n,m)$ by $n$ participants, and the last $n-m$ bits are randomly sampled permutations of the rest. By doing so, there are two benefits: 1) it avoids unsatisfactory SVs estimations, and 2) it improves convergence rate. Lines 11-17 show the within-round truncation operation at the sequence level. The idea is that successive sub-model evaluations in a sequence can be dismissed when the remaining marginal gain is smaller than a pre-defined truncation threshold $\epsilon_i$. Otherwise, the sub-model consisting of updates from $C$ participants is assembled and evaluated as shown in Lines 12-14. Then, participants' SVs is updated by their marginal gain in $\pi^k$.

\begin{algorithm}[t!]
\caption{FL Participant Contribution Evaluation}
\label{alg:fl}
\begin{flushleft}
\textbf{Input}: initial FL model $M^{(0)}$, evaluation function $V$, $n$ participants\\
\textbf{Output}: SVs $\phi_i, \forall i\in\{1,\dots,n\}$
\end{flushleft}
\begin{algorithmic}[1] 
\STATE $\phi_i=0 ,\forall i\in\{1,\dots,n\}$;
\FOR{$t=0,\dots,T-1$}
    \STATE \textit{\# run on client side};
    \FOR{each participant $i$ in parallel}
        \STATE $\Delta^{(t+1)}_i=$ ClientUpdate$(i,M^{(t)})$;
    \ENDFOR
    \STATE \textit{\# run on server side};
    \STATE $M^{(t+1)}=$ FedAvg$(\{\Delta^{(t+1)}_i\},M^{(t)})$;
    \STATE $\{\phi_i^{(t+1)}\}=$\AlgName$(M^{(t)},M^{(t+1)},V,\{\Delta_i^{(t+1)}\})$;
    \STATE $\phi_i=\phi_i+\phi_i^{(t+1)}, \forall; i\in\{1,\dots,n\}$;
\ENDFOR
\RETURN $\{\phi_{1},\dots,\phi_{n}\}$;
\end{algorithmic}
\end{algorithm}

The pseudo code for FL participant contribution evaluation using GTG-Shapley is shown in Algorithm \ref{alg:fl}. Line 2 shows that the FL training process lasts $T$ rounds. In each round $t$, participants download the global FL model $M^{(t)}$ to the client side. Line 5 indicates that participant $i$ trains the local model $M^{(t)}$ with its local datasets, and sends the gradient updates $\Delta^{(t+1)}_i$ to the FL server. Then, participants wait for the next round FL training. For FL server side (Line 8), the FL server collects the uploaded gradient updates $\{\Delta^{(t+1)}_i\}$ and executes FedAvg \cite{pmlr-v54-mcmahan17a} to update the FL model $M^{(t+1)}$. In Line 9, the FL server executes GTG-Shapley to compute the contributions by each participant in this round of training. For the $t$-th round, GTG-Shapley uses the previous FL model $M^{(t)}$, the updated FL model $M^{(t+1)}$, the evaluation function $V$ and the gradient updates from participants $\{\Delta^{(t+1)}_i\},\forall i\in\{1,\dots,n\}$ to estimate the latest SVs $\{\phi^{(t+1)}_{i}\},\forall i\in\{1,\dots,n\}$.

\subsection{Complexity Analysis}
The complexity of stand-alone Monte-Carlo approximation of Shapley Value has been studied in \cite{maleki2013bounding}. For each round of Shapley value computation, the Hoeffding's bound gives the minimal number of permutations needed for Monte-Carlo sampling to achieve an $(\epsilon,\delta)$-approximation, which is $K=(2r^2N/\epsilon^2)\log(2N/\delta)$, where $r$ is the range of marginal contributions. Therefore, the number of evaluations required over $T$ rounds of SV computation is expressed as $K_{total}=TK=\mathcal{O}(TN\log N)$.

GTG-Shapley performs within-round and between round truncation to accelerate SV computation. Similar as traditional machine learning, HFL also shows patterns of diminishing returns both on the marginal gain of within-round participant evaluation order in a sampled permutation (Figure \ref{fig:marginal}), and the marginal gain values across different rounds (Figure \ref{fig:g-SV}(b)). A participant can be categorized as `valuable' or `not valuable' based on its marginal gain. GTG-Shapley only needs to evaluate those `valuable' participants in a sampled permutation. For simplicity, we use bounded $Patero(\alpha_i,1,N)$ to model the probability of participants to be `valuable' at each position in a given sequence. The mean of the Patero distribution, $\overline{L_i}$, is the average length of participant combinations with valuable positions which GTG-Shapley needs to evaluate:
\begin{equation}
    \overline{L_i}=\frac{n^{\alpha_i}-n}{n^{\alpha_i}-1} \cdot \frac{\alpha_i}{\alpha_i-1}.
\label{eq:patero-i}
\end{equation}

By the same logic, the probability of the marginal gain of a given round categorized as `valuable' in all $T$ position can also be modeled as another bounded $Patero(\alpha_b,1,T)$. The average length $\overline{L_b}$ of `valuable' rounds in all $T$ rounds is:
\begin{equation}
    \overline{L_b}=\frac{T^{\alpha_b}-T}{T^{\alpha_b}-1} \cdot \frac{\alpha_b}{\alpha_b-1}.
\label{eq:patero-b}
\end{equation}

The number of evaluations required for GTG-Shapley to achieve an $(\epsilon,\delta)$-approximation is:
\begin{equation}
    K'_{total} = \overline{L_b} \cdot \overline{L_i} \cdot \frac{2(r-\epsilon')^2}{(\epsilon-\epsilon')^2} \log(\frac{2N}{\delta})
\end{equation}
where $\epsilon'=\max\{\epsilon_i,\epsilon_b\}$.

Depending on the data distribution, the complexity of GTG-Shapley is between $\mathcal{O}(T\log N)$ to $\mathcal{O}(TN\log N)$. For HFL involving i.i.d. datasets with an easy learning task, it is possible that the truncation point is right after the first position in a sequence, meaning that the ones after the first participant can hardly improve the FL model, and truncating them has little impact on Shapley value approximation. In such cases, $\overline{L_i}\approx 1$, meaning that the complexity of GTG-Shapley is $\mathcal{O}(T\log N)$. 

\section{Experimental Evaluation}
In this section, we compare GTG-Shapley with six state of the art existing approaches under various data distributions, and performance ablation studies on different variants of GTG-Shapley.

\subsection{Experiment Settings}
In this section, we first introduce the experiment settings and baselines. Then, we compare GTG-Shapley with the state of the art baselines under various FL settings involving both i.i.d. and non-i.i.d data silos.\footnote{Link to dataset and code is available in \url{ https://github.com/liuzelei13/GTG-Shapley}.}

\subsubsection{Datasets}
We adopt the FL data silos settings in \cite{wei2020efficient}. The dataset used in the experiments is based on the MNIST dataset \cite{lecun2010mnist}. It contains handwritten digit images. We randomly select 5,421 samples for each digit (i.e. 54,210 samples in total). Then, we randomly sample 892 images for each digit to be used as the test dataset (i.e. 8,920 samples in total). To evaluate the proposed algorithms under different FL settings, we design i.i.d. and non-i.i.d. FL scenarios with 10 participants as follows:
\begin{enumerate}
    \item \textbf{Same Distribution and Same Size}: We randomly sample 10,840 images and ensure that all 10 participants possess the same number of images (i.e. 1,084) for each digit.
    \item \textbf{Different Distributions and Same Size}: Each participant has the same number of samples. However, participant 1 \& 2's datasets contain 80\% of digits '1' and '2'. The other digits evenly divide the remaining 20\% of the samples. Similar procedures are applied to the rest of the participants.
    \item \textbf{Same Distribution and Different Sizes}: We randomly sample from the entire training set according to pre-defined ratios to form the local dataset for each participant, while ensuring that there are the same number of images for each digit in each participant. The ratios are 10\% for participants 1 and 2; 15\% for participants 3 and 4; 20\% for participants 5 and 6; 25\% for participants 7 and 8; and 30\% for participants 9 and 10.
    \item \textbf{Noisy Labels and Same Size}: We adopt the dataset from the Same Distribution and Different Sizes setting. Then, we flip the labels of a pre-defined percentage of samples in each participant's local dataset. The settings are: 0\% for participants 1 and 2; 5\% for participants 3 and 4; 10\% for participants 5 and 6; 15\% for participants 7 and 8; and 20\% for participants 9 and 10.
    \item \textbf{Noisy Features and Same Size}: We adopt the dataset from the Same Distribution and Same Size setting. Then, we add different percentages of Gaussian noise into the input images. The settings are: 0\% for participants 1 and 2; 5\% for participants 3 and 4; 10\% for participants 5 and 6; 15\% for participants 7 and 8; and 20\% for participants 9 and 10.
\end{enumerate}

\subsubsection{Comparison Approaches}
We compare GTG-Shapley against the following six approaches:
\begin{enumerate}
    \item \textbf{Original Shapley}: This method follows the principle of original SV calculation according to Eq. \eqref{eq:sv-og}. All possible combinations of participants are evaluated. Each sub-model is retrained from the dataset. The SV is calculated based on the model performance.
    \item \textbf{TMC Shapley}: Under this approach \cite{ghorbani2019data}, the model of a subset of FL participants is trained with their local datasets and the initial FL model. Monte-Carlo estimation of SVs is performed by random sampling permutations and truncating unnecessary sub-model training and evaluations.
    \item \textbf{Group Testing}: This method \cite{jia2019towards} samples a number of subsets of FL updates and evaluates the corresponding trained sub-model performance. Then, it estimates the Shapley differences instead of SVs. Afterward, SVs are derived from the Shapley differences by solving a feasibility problem. 
    \item \textbf{MR}: Under this approach \cite{song2019profit}, the model of a subset of FL participants is reconstructed based on each round of their gradient updates. The SV of each participant is calculated in each round. The final SV for a participant is the sum of his SVs in all rounds.
    \item \textbf{Fed-SV}: This method \cite{wang2020principled} approximates the 'federated Shapley value' by group testing-based estimation as GroupTesting. The differences are 1) the subsets performances used for estimating Shapley differences is evaluated on sub-model reconstructed by participants model parameters, and 2) SVs are estimated independently each round and aggregated later.
    \item \textbf{TMR}: Under this method \cite{wei2020efficient}, SVs are calculated independently in each round with participants' gradient updates, with a decay parameter $\lambda$ which serves as 1) weights to amplify SVs from earlier rounds, and 2) as a truncation factor to eliminate unnecessary sub-model re-constructions. 
\end{enumerate}
The convergence criterion for all approaches is set as:
\begin{equation}
\frac{1}{n*m}\sum^{10}_{m=1}{\sum^{n}_{i=1}{\frac{|\phi^k_i-\phi^{k-m}_i|}{|\phi^k_i|}}}<0.05.\label{eq:conv}
\end{equation}

\subsubsection{Performance Evaluation Metrics}
The performance of the comparison approaches are evaluated with the following metrics:
\begin{enumerate}
    \item \textbf{Time}: The total time of calculating SVs is used to evaluate the efficiency of each approach. In order to show different baselines in the same perspective, we apply $\log_{10}(\cdot)$ on the total time elapsed.
    \item \textbf{Cosine Distance}: We denote the SV result calculated by Original Shapley as a vector $\phi^*=<\phi^*_1,\dots,\phi^*_n>$, and the estimation result calculated by any other approach as $\phi=<\phi_1,\dots,\phi_n>$. The Cosine Distance is defined as: 
    \begin{equation}
    1-cos(\phi^*,\phi). 
    \end{equation}
    \item \textbf{Euclidean Distance}: The Euclidean Distance is defined as:
    \begin{equation}
    \sqrt{\sum^n_{i=1}{(\phi^*_i-\phi_i)^2}}. 
    \end{equation}
    \item \textbf{Maximum Difference}: The Maximum Difference is defined as:
    \begin{equation}
    {\max}^n_{i=1}|\phi^*_i-\phi_i|. 
    \end{equation}
\end{enumerate}
The smaller the values achieved for all four metrics, the better the performance of an approach.

\subsection{Results and Discussions}
\begin{figure}[tb!]
    \centering
    \includegraphics[width=0.5\linewidth]{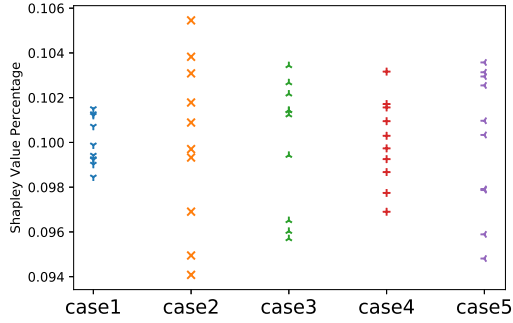}
    \caption{The original Shapley Values under the 5 experiment settings.}
    \label{fig:og-sv}
\end{figure}

In this section, we analyze the experimental results under the five aforementioned settings. The original Shapley Values for the five settings are shown in Figure \ref{fig:og-sv}. In each case ,there are 10 points representing participants' contributions. In the following subsection, both the x-axis and the y-axis for each figure are presented in $\log$ scale.

\subsubsection{Same Distribution and Same Size} 
\begin{figure}[t!]
    \centering
    \includegraphics[width=0.95\linewidth]{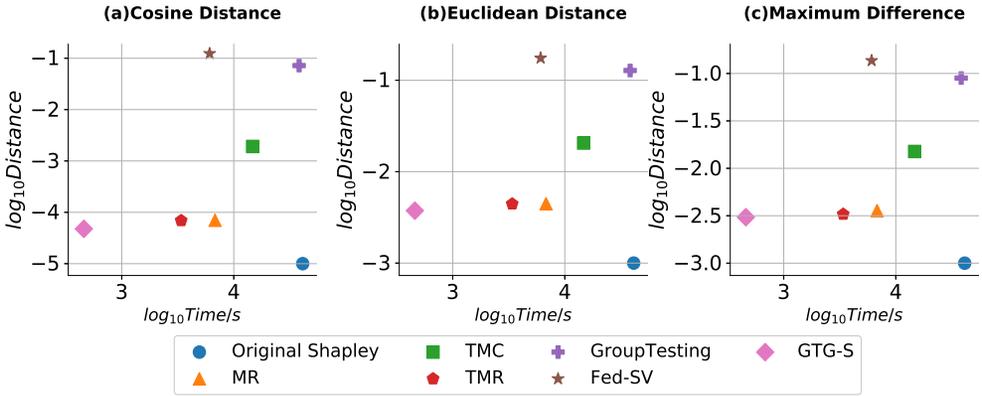}
    \caption{SV Convergence with respect to Time under Same Distribution and Same Size}
    \label{fig:result1}
\end{figure}

As shown in Figure \ref{fig:result1}, the dataset based baselines are much slower than the gradient-based approaches. In terms of efficiency, MR, TMR and Fed-SV show similar results and are much faster than GroupTesting and Original Shapley. GTG-Shapley achieves the highest efficiency, significantly faster than the best performing baseline TMR by 7.4 times. In terms of accuracy, all three distance metrics show that GTG-Shapley, MR and TMR has the similar results, with GTG-Shapley slightly outperforming MR and TMR.

\begin{figure}[t!]
    \centering
    \includegraphics[width=0.95\linewidth]{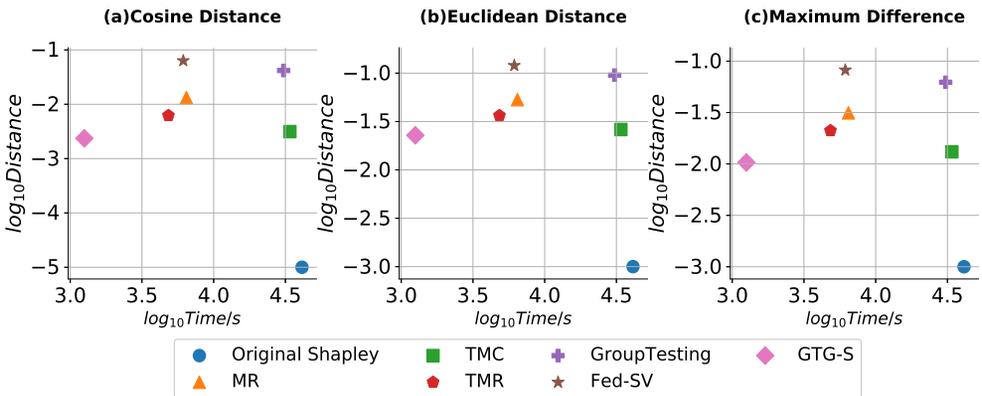}
    \caption{SV Convergence with respect to Time under Different Distributions and Same Size}
    \label{fig:result2}
\end{figure}

\begin{figure}[t!]
    \centering
    \includegraphics[width=0.95\linewidth]{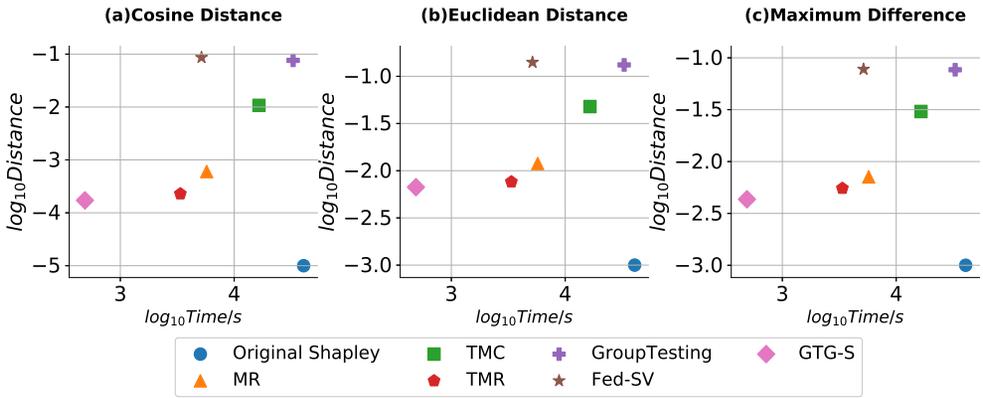}
    \caption{SV Convergence with respect to Time under Same Distribution and Different Sizes}
    \label{fig:result3}
\end{figure}

\begin{figure}[t!]
    \centering
    \includegraphics[width=0.95\linewidth]{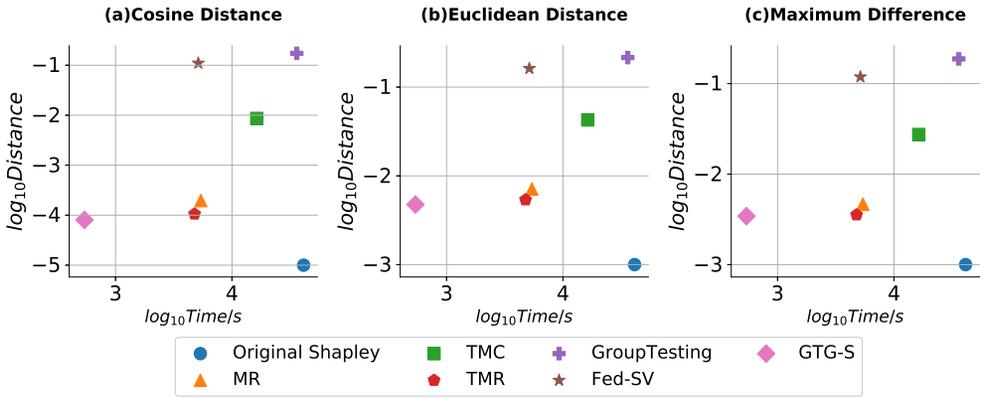}
    \caption{SV Convergence with respect to Time under Noisy Labels and Same Size}
    \label{fig:result4}
\end{figure}

\begin{figure}[t!]
    \centering
    \includegraphics[width=0.95\linewidth]{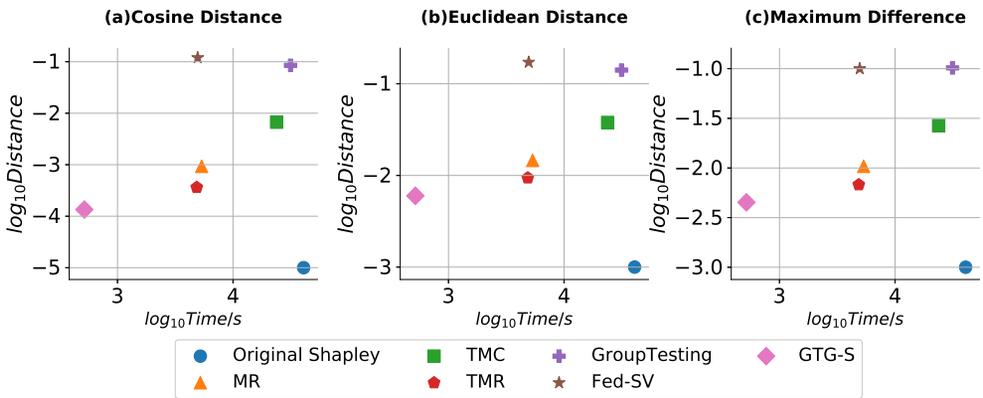}
    \caption{SV Convergence with respect to Time under Noisy Features and Same Size}
    \label{fig:result5}
\end{figure}

\subsubsection{Different Distributions and Same Size} 
In terms of efficiency, Figure \ref{fig:result2} shows a similar pattern as in Figure \ref{fig:result1}. GTG-Shapley is the fastest approach, with TMR trailing behind. On the other hand, in terms of accuracy, GTG-Shapley achieves the best accuracy under all three distance metrics. The accuracy gap is larger than that under the condition in Figure \ref{fig:result1}. TMC surpasses MR and TMR as the second most accurate, and comes close to GTG-Shapley. It shows that GTG-Shapley is well suited for this non-i.i.d. setting.

\subsubsection{Same Distribution and Different Sizes} 
Under this setting (Figure \ref{fig:result3}), GTG-Shapley maintains its superior efficiency over other approaches under this non-i.i.d. setting. MR and Fed-SV achieves similar efficiency, but are slower than TMR. In term of accuracy, GTG-Shapley outperforms others in this scenarios as well and with the distance from the actual SVs remaining below $1\times 10^{-2}$ for all three distance metrics. This is followed by TMR which also has a noticeable lead on other baselines.

\subsubsection{Noisy Labels and Same Size} 
As shown in Figure \ref{fig:result4},  the results indicate that GTG-Shapley is significantly faster than others under this non-i.i.d. setting. Under this scenario, Fed-SV, MR and TMR achieve similar efficiency. TMC falls in the middle between gradient-based MR and other dataset based methods. In terms of accuracy, GTG-Shapley achieves the best results among all baselines. TMR and MR follow GTG-Shapley closely, and lead the next best approach, TMC, with a clear gap.

\subsubsection{Noisy Features and Same Size} 
From Figure \ref{fig:result5}, it can be observed that GTG-Shapley achieves significantly higher efficiency over all other approaches under this non-i.i.d. setting. TMR shows similar efficiency to MR as TMC to Original Shapley. In terms of accuracy, GTG-Shapley achieves the best results and it leads TMR and MR with a small gap. TMC is in the middle between MR and GroupTesting.

In summary, GTG-Shapley achieves the best performance in efficiency and accuracy of estimating SVs under i.i.d. and non-i.i.d. settings. It is consistently and significantly faster than the best performing baseline, and either matches or slightly outperforms the latter in terms of accuracy. General observations can be made that gradient-based methods are significantly faster over dataset based methods. Truncation techniques can improve efficiency. However improvement from truncation is sensitive to non-i.i.d. settings (e.g., TMC slows down in scenarios with different distributions and with noisy features. TMR slows down in scenarios with noisy labels and features.) In all scenarios, GTG-Shapley shows consistent efficiency improvement even when other methods with truncation techniques are negatively affected. This can be attributed to the partial permutation sampling design in GTG-Shapley, which improves SV estimation convergence.

In terms of accuracy, GTG-Shapley achieves the best results in all scenarios. The SV estimates it computed remains below $1\times 10^{-2}$ away from the actual Shapley values in all scenarios except Scenarios 2. General observation can be made that multi-stage calculation of SVs achieves better accuracy in most scenarios except Scenario 2. Moreover, truncation techniques are more robust against non-i.i.d. settings.


\subsection{Ablation Study}
In this section, we analyze experimentally the effect of the main components of GTG-Shapley. The experimental settings are the same as in the previous experimental evaluation section.

\subsubsection{GTG-Shapley Variants}
We compare the full version of GTG-Shapley against the following variants of GTG-Shapley.
\begin{enumerate}
    \item \textbf{GTG-Ti}: This approach contains only the within-round truncation component in GTG-Shapley. The between round truncation and guided sampling components of GTG-Shapley are excluded.
    \item \textbf{GTG-Tib}: This approach contains the within-round and between round truncation components in GTG-Shapley, but does not contain the guided sampling component.
    \item \textbf{GTG-OTi}: This approach is used for analysing the effect of the frequency of computing SVs. GTG-OTi accumulates the gradients from every FL participant from the beginning, and performs gradient-based SV estimation only once after the FL training is complete. The within-round truncation is performed when computing the SVs.
\end{enumerate}
The convergence criterion  for all approaches is the same as in Eq. \eqref{eq:conv}.

\begin{figure}[b]
    \centering
    \includegraphics[width=0.95\linewidth]{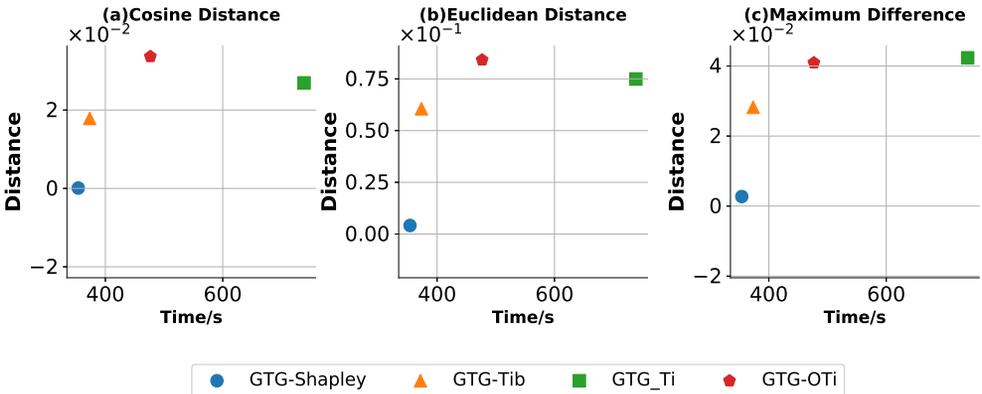}
    \caption{Same Distribution and Same Size}
    \label{fig:result1-g}
\end{figure}

\begin{figure}[t]
    \centering
    \includegraphics[width=0.95\linewidth]{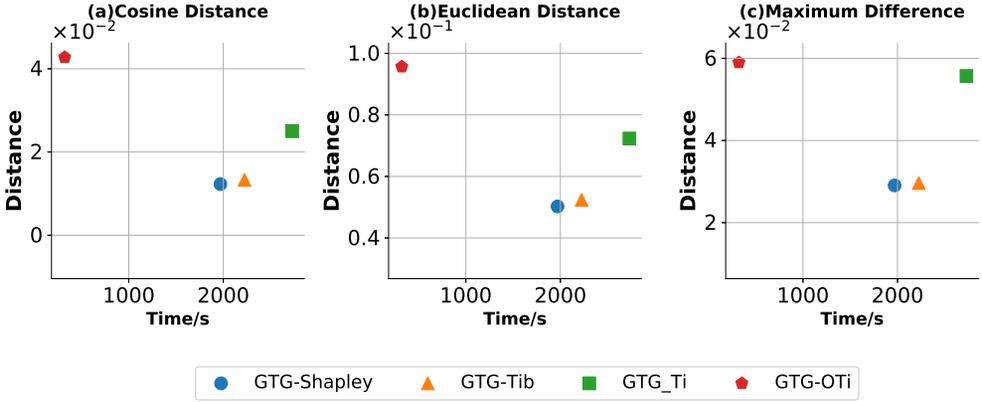}
    \caption{Different Distributions and Same Size}
    \label{fig:result2-g}
\end{figure}

\begin{figure}[t]
    \centering
    \includegraphics[width=0.95\linewidth]{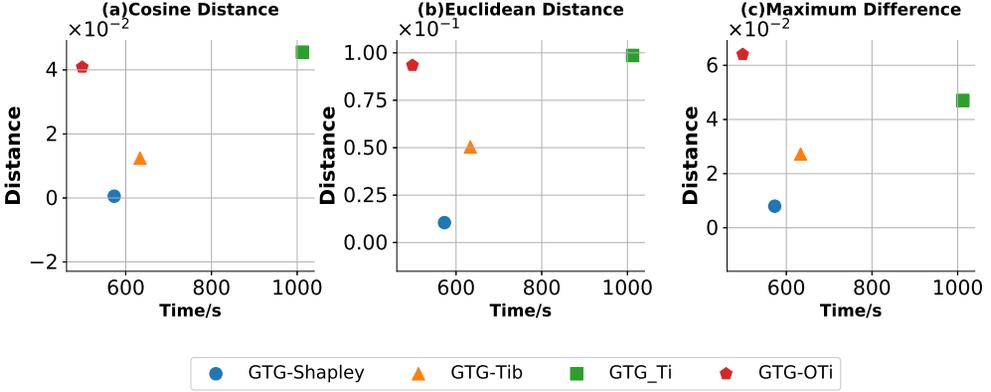}
    \caption{Same Distribution and Different Sizes}
    \label{fig:result3-g}
\end{figure}

\begin{figure}[t]
    \centering
    \includegraphics[width=0.95\linewidth]{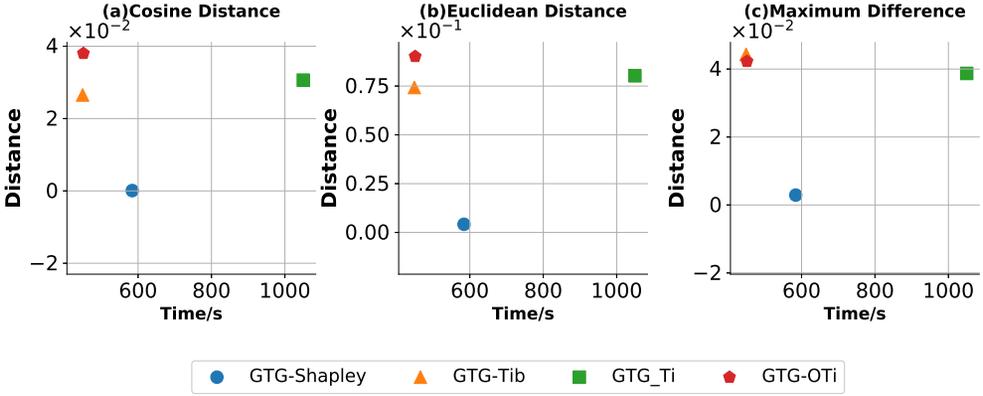}
    \caption{Noisy Labels and Same Size}
    \label{fig:result4-g}
\end{figure}

\begin{figure}[t]
    \centering
    \includegraphics[width=0.95\linewidth]{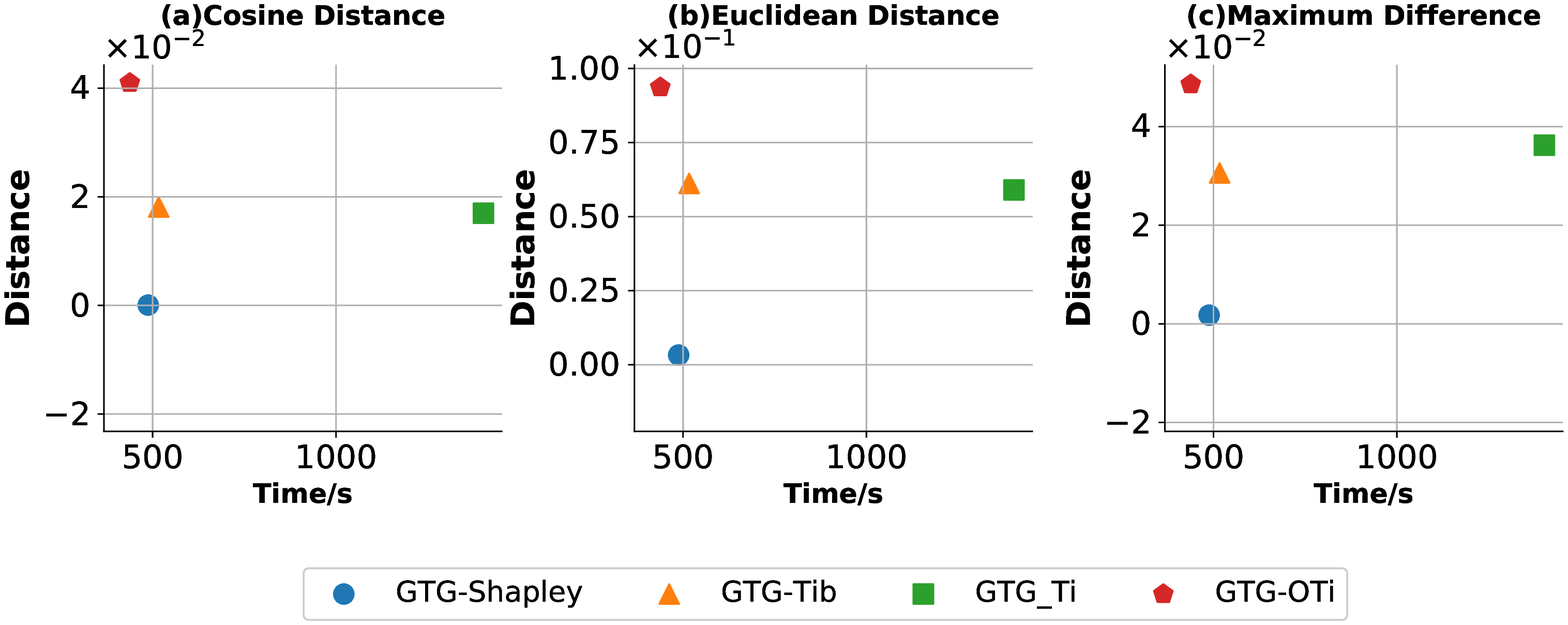}
    \caption{Noisy Features and Same Size}
    \label{fig:result5-g}
\end{figure}

\subsubsection{The Effect of SV Computing Frequency}
Inspired by the design different computing frequency of MR and OR in \cite{jia2019towards}, we study the effect of SV computing frequency on GTG-Shapley in terms of efficiency and accuracy. GTG-Ti and GTG-OTi are basically the same except in terms of SV computing frequency. GTG-Ti has the highest frequency (i.e. in every round) and GTG-OTi has the lowest frequency (i.e. only once in total). As shown on all 5 settings in Figures \ref{fig:result1-g} to \ref{fig:result5-g}, GTG-OTi maintains a 2~3 time lead in terms of efficiency compared to GTG-Ti. 
The reason is that the overall marginal utility value drops across different rounds (Figure \ref{fig:marginal}). Thus, GTG-Ti is most likely to truncate more sub-model evaluations in later rounds. In addition, in terms of efficiency, it is worth noting that GTG-Shapley and GTG-Tib are slightly faster than GTG-OTi in i.i.d settings, as shown in Figure \ref{fig:result1-g}. Considering GTG-Shapley and GTG-Tid both need to compute SVs in every round, the reasons behind the faster computing is most likely due to: 1) the i.i.d. settings are more conducive for FL models to converge so that most of the marginal utility values are gained during the beginning rounds, which reduces GTG-OTi's advantage in low SV computing frequencies; and 2) GTG-OTi directly utilizes the gradient updates  accumulated over multiple rounds to reconstruct FL sub-models. This may result in utilities deviating from the ground truth further, which negatively affects SV estimation convergence. The negative impact may not be so significant such that GTG-OTi is only slower than the two approaches in i.i.d settings. In all four Non-i.i.d settings, it is always faster than the other three approaches as shown in Figure \ref{fig:result2-g}.

GTG-OTi's high efficiency is achieved at the cost of lower accuracy. It performs the worst among all four approaches, as shown in Figures \ref{fig:result1-g} to \ref{fig:result5-g}. For Monte-Carlo estimation, the deviation from GTG-OTi's low SV computing frequency significantly influences the utility value of the respective reconstructed FL sub-models, which leads to poorer performance comparing to others. 
Thus, reducing SV computing frequency can improve SV estimation efficiency, but with a negative impact on the accuracy of the estimated SVs.

\subsubsection{The Effect of Between-Round Truncation}
Here, we focus on comparing GTG-Tid and GTG-Ti to study the effect of between-round truncation. Since the only difference between the two is that GTG-Tid includes between-round truncation and GTG-Ti does not. GTG-Tid is 17\% more efficient than GTG-Ti (Figure \ref{fig:result2-g}), and can be up to 3 times faster shown in Figure \ref{fig:result5-g}. The efficiency improvement depends on the FL model convergence rate. The between-round truncation technique can further improve efficiency under i.i.d settings (Figure \ref{fig:result1-g}).

As for the effect on accuracy, between-round truncation can slightly improve SVs estimation accuracy under some data distribution settings. GTG-Tid achieves higher accuracy than GTG-Ti in Figures \ref{fig:result1-g}, \ref{fig:result2-g} and \ref{fig:result3-g}. Under other data distribution settings in Figures \ref{fig:result4-g} and \ref{fig:result5-g}, the two approaches achieve similar accuracy performance.
Thus, between-round truncation can improve the SV estimation efficiency while increasing or maintaining the accuracy of the estimated SVs.

\subsubsection{The Effect of Guided Sampling}
Here, we focus on comparing GTG-Shapley and GTG-Tid to study the effect of guided sampling. The main purpose of the proposed guided sampling technique is to improve accuracy with participant orders which allow different participants to occupy important positions in different permutation samplings to reduce possible bias in SV estimation. With guided sampling, GTG-Shapley achieves the highest accuracy in all settings, as shown in Figures \ref{fig:result1-g} to \ref{fig:result5-g}. GTG-Shapley achieves orders of magnitude higher accuracy than others in most cases.

As for the effect on efficiency, GTG-Shapley also outperforms GTG-Tid in most cases. Thus, guided sampling can significantly improve the SV estimation accuracy while enhancing or maintaining efficiency. The ablation study shows that all three design considerations in the proposed GTG-Shapley approach are important in improving the efficiency and accuracy of estimating SVs.


\section{Conclusions and Future Work}
In this paper, we proposed an efficient approach to estimate Shapley values, the Guided Truncation Gradient Shapley (\AlgName), to address the limitation of existing FL participant contribution evaluation approaches. It can evaluate participants' Shapley value-based contributions to the FL model performance without requiring extra model training or exposing sensitive local data. In addition, it utilizes previously evaluated model utilities as a guidance to opportunistically skip unnecessary sub-model reconstruction and evaluation operations, which further improves efficiency. 
Extensive experimental evaluation demonstrates that GTG-Shapley can closely approximate the actual Shapley values, while significantly improving computation efficiency compared to state-of-the-art approaches.

In subsequent research, we will focus on building FL participant auditing tools based on GTG-Shapley in order to support regulatory bodies to provide oversight to federated learning-based data exchange.

\begin{acks}
This research is supported by the National Research Foundation, Singapore under its AI Singapore Programme (AISG Award No: AISG2-RP-2020-019); the Joint NTU-WeBank Research Centre on Fintech (Award No: NWJ-2020-008), Nanyang Technological University, Singapore; the Nanyang Assistant Professorship (NAP); the RIE 2020 Advanced Manufacturing and Engineering (AME) Programmatic Fund (No. A20G8b0102), Singapore; and the Joint SDU-NTU Research Centre on Artificial Intelligence (C-FAIR), Shandong University, China (NSC-2019-011); the National Key Research and Development Program of China under Grant (No. 2018AAA0101100).
\end{acks}

\bibliographystyle{ACM-Reference-Format}
\bibliography{gtg-shapley}

\end{document}